\documentclass{article}
\usepackage{
    amsmath,
    amssymb,
    bm,
    booktabs,
    graphicx,
    hyperref,
    makecell,
    multirow,
    siunitx,
    spconf,
    tabularx,
    todonotes,
    cite
    }

\newcolumntype{L}{>{\raggedright\arraybackslash}X}   
\newcolumntype{R}{>{\raggedleft\arraybackslash}X}

\DeclareMathOperator*{\argmin}{arg\,min}

\newcommand{\DEAP}{\textit{DEAP}}
\newcommand{\DREAMER}{\textit{DREAMER}}
\newcommand{\SEED}{\textit{SEED}}
\newcommand{\SEEDIV}{\textit{SEED-IV}}

\newcommand{\Negative}{\textit{Negative}}
\newcommand{\Neutral}{\textit{Neutral}}
\newcommand{\Positive}{\textit{Positive}}

\title{Domain-Invariant Representation Learning from EEG with\\Private Encoders}

\makeatletter
\def\@name{ \emph{David Bethge$^{1,2}$, Philipp Hallgarten$^{1,3}$, Tobias Grosse-Puppendahl$^{1}$, Mohamed Kari$^{1}$},  \\ \emph{\textit{Ralf Mikut}$^{3}$, \textit{Albrecht Schmidt}$^{2}$, \textit{Ozan \"{O}zdenizci}$^{4,5}$} \\}
\makeatother

\address{
$^{1}$ Dr. Ing. h.c. F. Porsche AG, Stuttgart, Germany \\
$^{2}$ Ludwig-Maximilians University Munich, Germany \\
$^{3}$ Karlsruhe Institute of Technology, Germany \\
$^{4}$ Institute of Theoretical Computer Science, Graz University of Technology, Austria \\
$^{5}$ TU Graz - SAL Dependable Embedded Systems Lab, Silicon Austria Labs, Austria
}
\begin{document}



\maketitle
\begin{abstract}
Deep learning based electroencephalography (EEG) signal processing methods are known to suffer from poor test-time generalization due to the changes in data distribution.
This becomes a more challenging problem when privacy-preserving representation learning is of interest such as in clinical settings.
To that end, we propose a multi-source learning architecture where we extract domain-invariant representations from dataset-specific private encoders.
Our model utilizes a maximum-mean-discrepancy (MMD) based domain alignment approach to impose domain-invariance for encoded representations, which outperforms state-of-the-art approaches in EEG-based emotion classification.
Furthermore, representations learned in our pipeline preserve domain privacy as dataset-specific private encoding alleviates the need for conventional, centralized EEG-based deep neural network training approaches with shared parameters.
\end{abstract}
\section{Introduction}

Over the past decades, electroencephalography (EEG) based emotion recognition gained significant interest towards developing affective neural-machine interfaces \cite{muhl2014survey}. 
However, due to the costly data collection processes, large EEG datasets recorded under emotion eliciting experimental paradigms are not ubiquitous.
Several small scale affective EEG datasets were previously introduced, which however highly differ in their experimental setups (\textit{e.g.}, stimuli, emotion labels).
This introduces the well-known \textit{domain adaptation} problem, i.e., models trained to recognize emotions on one of these datasets commonly fails to solve the same task for another dataset.
In order to make EEG-based emotion recognition algorithms suitable for a variety of experiments and real-life scenarios, it is crucial to obtain a model that can tackle this task across multiple datasets.
One approach to achieve this from a deep learning perspective is to extract and exploit domain-invariant representations from multi-channel EEG data.

Recent work have shown significant promise in using invariant representation learning from time-series EEG data for cross-subject generalization \cite{ozdenizci2020learning,jeon2021mutual,rayatdoost2021subject}, while learning across multiple dataset sources remains an open question.
Notable methods successfully introduced adversarial censoring for domain invariance, widely known as DANN \cite{ganin2016domain}, in various EEG decoding tasks \cite{ozdenizci2020learning,jeon2021mutual}, including emotion recognition \cite{rayatdoost2021subject}. 
Adversarial domain regularization with DANN \cite{ganin2016domain} considers a cross-domain shared encoder that extracts features from data of all subjects, and two classifiers: a task and a domain classifier. 
During training the encoder is adversarially penalized on the domain classification loss, which enforces the model towards learning domain-invariant representations. 
In this work we take a different approach to invariant EEG representation learning by further considering to preserve domain privacy that is of critical importance in clinical settings \cite{agarwal2019protecting,ju2020federated}.

We propose a multi-source learning framework for domain invariant representation learning from time-series signals such as multi-channel EEG recordings. 
From a different perspective than adversarial learning methods, our framework consists of a private feature encoder per domain and a cross-domain shared classifier, where we utilize a maximum-mean-discrepancy (MMD)~\cite{gretton2012kernel} based domain alignment loss across private feature encoders to minimize domain-specific leakage within the learned representations.
Our contributions in this work are three-fold:
(1) We introduce a deep neural signal processing architecture where EEG time-series are privately encoded at the source end and only the learned representations are shared to a global classifier that enables decentralized cross-dataset learning by preserving domain privacy.
(2) We reveal large dataset domain specific variances in conventionally trained centralized pipelines, and demonstrate that regularizing latent representations via an MMD-based domain alignment loss enables dataset source independent representation learning.
(3) We show that the use of adaptive batch normalization layers in such multi-source settings prevent performance decrease at inference time.

\section{Materials and Methods}

\subsection{Experimental Data}

\textbf{Benchmark Affective EEG Datasets:} We used four publicly available EEG-based emotion recognition datasets: \DEAP\ ~\cite{koelstra2011deap}, \DREAMER\ ~\cite{katsigiannis2017dreamer}, \SEED\ ~\cite{duan2013differential, zheng2015investigating}, \SEEDIV\ ~\cite{zheng2018emotionmeter}. 
During EEG recordings subjects were presented audio-visual stimuli that are expected to elicit distinct emotions.
Differences in the experimental setups lead to a variability in the structure of data samples and their labels.
\textit{SEED} and \textit{SEED-IV} contain $62$-channel EEG recordings sampled at \SI{200}{\hertz}, whereas \textit{DEAP} contains recordings from $32$ electrodes and \DREAMER\ from $14$, both sampled at \SI{128}{\hertz}.
Emotion labels of \SEED\ and \SEEDIV\ represents one out of three and four discrete emotions respectively, labeled by the stimuli that the subjects were presented.
For \DEAP\ and \DREAMER\, post-recording subject self-assessment ratings on the \textit{valence} and \textit{arousal} continuous scales were included as the labels.

\vspace{0.1cm}
\noindent\textbf{Label Transformation:} To realize our cross-dataset experimental analyses, we transformed the label spaces across all four datasets into a common set. We determined the three discrete emotions \Negative, \Neutral, and \Positive\ (as also used in the \SEED-dataset) as the common label-space. Using \textit{k-means} clustering in the two-dimensional label space of \textit{Valence} and \textit{Arousal} for \DEAP, and \DREAMER\ datasets, we determined four clusters corresponding to the four discrete emotions used as labels in the \SEEDIV\ dataset: \textit{Fear}, \textit{Sad}, \textit{Neutral}, \textit{Happy}. We merged the former two into being the \Negative\ class, and considered \textit{Happy} to be the \Positive\ class. Hence we obtained three class labels in a unified manner.
\begin{figure*}[th!]
    \centering
    \includegraphics[width=\textwidth]{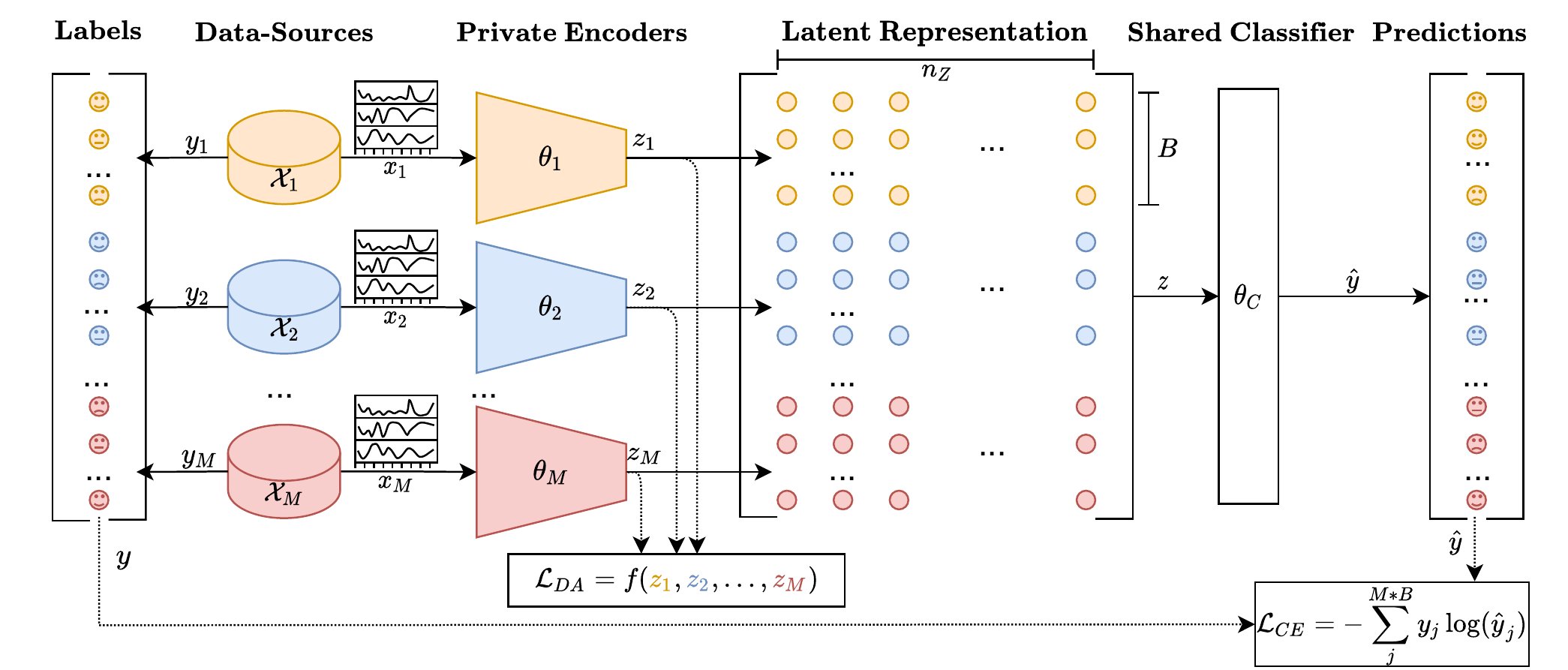}
    \caption{\textbf{DAPE Architecture.} Each color represents one data-source. Signals $x_1$ from each data-source $\mathcal{X}_i$ are processed by a private encoder. Output representations $z_i$ are then used to calculate the domain-alignment loss $\mathcal{L}_{DA}$ via Eq.~\eqref{eq:lda}. In addition, output representations are concatenated to a representation $z \in \mathbb{R}^{(B*M) \times n_z}$ and used as input for a shared classifier. Classifier outputs a batch of predictions $\hat y_i$ for each batch of  signals $x_i$. Corresponding batches of labels $y_i$ are concatenated in the same way as the latent representations $z_i$ and used together with the predictions for the calculation of the cross-entropy loss $\mathcal{L}_{CE}$.}\vspace{0.25cm}
    \label{fig:DAPE-Architecture}
\end{figure*}

\begin{table*}[tb]
    \caption{Accuracies of 3-class emotion classification evaluated with \textit{DAPE}, \textit{aDAPE} and baseline methods: local and global baseline models, and \textit{DANN} architecture. Bottom row shows the results of the linear SVM used to assess the 4-class domain-invariance. Local baseline corresponds to using four independent networks (i.e., dataset-specific encoder and classifiers) without any regularization or domain-alignment.}\vskip0.5em
    \label{tab:results}
    \centering
    \begin{tabularx}{\textwidth}{Xr|cc|ccc}
    \toprule
    & & Local Models & Global Model & \textit{DANN}~\cite{ganin2016domain} & \textit{DAPE} (\textbf{Ours}) & \textit{aDAPE} (\textbf{Ours}) \\
    \midrule
    \multirow{5}{*}{\makecell[l]{Emotion\\Classification ($\nearrow$)}}
    &\DEAP&$50.01\%$ {\footnotesize($\pm 0.58$)} &$40.02\%$ {\footnotesize($\pm 1.09$)}&$39.81\%$ {\footnotesize($\pm 0.87$)}&$47.81\%${\footnotesize($\pm 0.24$)}&$47.99\%${\footnotesize($\pm 1.30$)} \\
    &\DREAMER&$59.03\%$ {\footnotesize($\pm 2.41$)} &$43.78\%$ {\footnotesize($\pm 1.28$)}&$42.30\%$ {\footnotesize($\pm 1.60$)}&$49.13\%${\footnotesize($\pm 1.54$)}&$48.70\%${\footnotesize($\pm 1.99$)}\\
    &\SEED& $56.19\%$ {\footnotesize($\pm 1.97$)} &$42.93\%$ {\footnotesize($\pm 0.95$)}&$42.63\%$ {\footnotesize($\pm 0.26$)}&$51.72\%${\footnotesize($\pm 0.67$)}&   $53.45\%${\footnotesize($\pm 2.05$)}\\
    &\SEEDIV&$42.42\%$ {\footnotesize($\pm 3.64$)} &$34.55\%$ {\footnotesize($\pm 0.78$)}&$34.57\%$ {\footnotesize($\pm 0.56$)}&$41.94\%${\footnotesize($\pm 0.64$)}& $43.25\%${\footnotesize($\pm 1.07$)}\\
    &\textbf{Mean}&$51.91\%$ {\footnotesize($\pm 2.15$)} &$40.32\%$ {\footnotesize($\pm 0.42$)}&$39.82\%$ {\footnotesize($\pm 0.33$)}&$47.65\%${\footnotesize($\pm 0.13$)}& $\bm{49.09\%}${\footnotesize($\pm 1.49$)}\\
    \midrule
    \makecell[l]{Domain\\ Classification ($\searrow$)}
    &\textbf{Mean}  &$99.87\%$ {\footnotesize($\pm 0.15$)} &$60.11\%$ {\footnotesize($\pm 3.47$)}&$66.17\%${\footnotesize($\pm 4.45$)}&$82.83\%${\footnotesize($\pm 3.18$)}&$\bm{52.76\%}${\footnotesize($\pm 2.92$)}\\
    \bottomrule
    \end{tabularx}
\end{table*}

\subsection{Multi-Source EEG Processing for Private Encoding}

We will denote multiple datasets $\mathcal{D}_k$ consisting of time-series EEG signal epochs and corresponding emotion label of the experiment paradigm as pairs $(s_j,l_j)$.
We pre-process the signals to serve as input to our architecture as follows. 
First, we perform baseline correction for signals $s_j$ using a three-second time-window and apply a \SI{4}-\SI{40}{\hertz} Butterworth band-pass filter.
Since the length of the available $s_j$ differ within each dataset, we used only the last $T$ seconds of each available signal with $T$ being the length of the shortest time-series in the dataset, i.e., 60, 64, 185, 50 for \DEAP, \DREAMER, \SEED\ and \SEEDIV\ respectively.
Then, we segment the signals into non-overlapping \SI{2}{\second} windows that will be used as inputs $x_i$ to our model. 
We assign the label $l_{j_i}$ to each window $x_i$. 
All pairs of $x_i$ and corresponding labels $y_i$ obtained from a dataset are considered as a processed data-source $\mathcal{X}_k$. 

To balance the number of samples across data-sources and with respect to the label for each data-source individually, we applied stochastic undersampling while constructing our training set.
We split the data into the training ($60\%$), validation ($20\%$) and test ($20\%$) sets, while we assure that the stratification constraints hold for each subset individually.

\subsection{Multi-Source Learning Framework}

\textbf{Domain Aligned Private Encoders (DAPE):} Our model consist of one private encoder per data-source and a shared classifier.
While the encoder can be designed arbitrarily for each data-source, for proof of concept we used the \textit{DeepConvNet} architecture~\cite{schirrmeister2017deep} for our all data-sources. 
We replaced the final pooling layer in \textit{DeepConvNet} by an adaptive average pooling layer, that introduces the flexibility to control the number of features $n_z$ in the latent representation $z$.
In our models we chose $n_z=50$ as the best results in terms of the balance between emotion classification and domain invariance were achieved.
One extension of our model towards achieving better domain adaptation, which we denote by adaptive DAPE (aDAPE), considers the use of adaptive batch normalization layers~\cite{li2016revisiting}. Such layers were also recently used in deep EEG analysis by~\cite{jimenez2020custom,jimenez2021standardization}, instead of conventional batch normalization.
By doing so, we ensure each layer of any given encoder to receive data from a similar distribution at test time as well, as samples are normalized by the statistics estimated from the utilized mini-batch.

During each forward pass, a batch of $B$ samples from $x_i\sim\mathcal{X}_i$ are drawn and used as input to each of the $M$ encoders, represented through the parameters $\theta_i, i \in \{1,\ldots,M\}$, with $M$ also being the number of data-sources.
We consider the output representations $z_{i} = f(\theta_i, x_i) \in \mathbb{R}^{B \times n_z}$ of each encoder as samples drawn from independent data-source distributions.
To overcome the problem of domain shift across encoders and regularize the encoders towards learning domain-invariant representations, we use an MMD~\cite{gretton2012kernel} based domain-alignment loss $\mathcal{L}_{\text{DA}}$ using Gaussian kernels with $\sigma_j\in\{10, 15, 20, 50\}$. 
Rather than computing the MMD for all combinations of data-sources, we randomly sample $M$ pairs of distributions $\{(z_p^{(i)}, z_q^{(i)})\}_{i=1}^M$, $p,q \in \{1,\ldots,M\}$ such that $p \neq q$, calculate the MMD loss~\cite{gretton2012kernel} between them and accumulate in the domain alignment regularizer:
\begin{equation}\label{eq:lda}
    \mathcal{L}_{DA} = \sum_{j=1}^{\vert\sigma \vert}\sum_{i=1}^{M} \text{MMD}^2(z_{p}^{(i)}, z_{q}^{(i)}, \sigma_j),
\end{equation}
where $\vert\sigma\vert$ denotes the cardinality for the pre-defined set $\sigma_j$ is chosen from and $\text{MMD}^{2}(z_{p}^{(i)},z_{q}^{(i)},\sigma_j)$ is defined as follows:
\begin{align}
\text{MMD}^{2} =& \frac{1}{B(B-1)} \sum_{m=1}^B \sum_{\substack{n=1\\n\neq m}}^B \varphi(z_{p_m}^{(i)}, z_{p_n}^{(i)}, \sigma_j) \nonumber \\
& + \frac{1}{B(B-1)} \sum_{m=1}^B \sum_{\substack{n=1\\n\neq m}}^B \varphi(z_{q_m}^{(i)}, z_{q_n}^{(i)}, \sigma_j) \nonumber\\
& - \frac{2}{B^2} \sum_{m=1}^B \sum_{n=1}^B \varphi(z_{p_m}^{(i)}, z_{q_n}^{(i)}, \sigma_j) ,
\end{align}
with $\varphi(z_{p_m}^{(i)}, z_{q_n}^{(i)}, \sigma_j)$ being the Gaussian kernel function:
\begin{equation}
    \varphi(z_{p_m}^{(i)}, z_{q_n}^{(i)}, \sigma_j) = \exp \left(\frac{- \Vert z_{p_m}^{(i)} - z_{q_n}^{(i)} \Vert^{2}}{2\sigma_j^{2}}\right).
\end{equation}
All batches of representations $z_i$ are then stacked to one vector of representations $z\in\mathbb{R}^{(B*M) \times n_z}$, and used as input for the classifier with parameters $\theta_C$ to compute predicted emotions $\hat y$. 
We train the network with a categorical cross-entropy loss function $\mathcal{L}_{CE}= -\sum_j^{M*B} y_j \log(\hat y_j)$.
Note that only the parameters $\theta_C$ and $\theta_i$ (parameters of the encoder that processed a sample $x_j$) contribute to the prediction of an individual sample $\hat y_j$, which ensures that only the classifier and the encoder $i$ are optimized based on the gradient of the loss.

\vspace{0.2cm}
\noindent\textbf{Multi-Objective Optimization:} Training of \textit{DAPE} can be considered as a multi-objective optimization. 
On one hand, we want to minimize the cross-entropy loss for the classification task and one the other hand we want to minimize the domain-alignment loss $\mathcal{L}_{\text{DA}}$ for learning domain-invariant representations. 
We assume that these goals are not contradictory and that the classification task gets more robust using domain-invariant representations of the samples.
To control the domain-invariance of representations we use a hyperparameter $\kappa$ as the tunable weight in the optimization problem:
\begin{equation}
   \theta_1^{*}, \ldots, \theta_M^{*}, \theta_C^{*} = \argmin_{\theta_1, \ldots, \theta_M, \theta_C}{\mathcal{L}_{CE} + \kappa \mathcal{L}_{DA}}.
\end{equation}
We use an annealing update scheme for $\kappa$ as proposed by~\cite{schonfeld2019generalized,bowman2015generating}.
To ensure that the domain alignment does not lead the encoders to learn task-irrelevant representations, we start the training with $\kappa=0$ and increase it with the beginning of the 5th epoch by a rate of $0.25$ per epoch. 
To prevent the domain-alignment loss dominating the cross-entropy loss in the later epochs, we stop updating $\kappa$ by the 70th epoch and retain the value of $16.25$ for all subsequent epochs.

\subsection{Quantifying Domain-Invariance of Representations}

One of the motivating goals in our architecture is to preserve domain privacy at the data-source end, as well as ensuring that an attacker not being able to deduce the information concerning the data-source from latent representations. 
To assess invariance of learned representations via DAPE, we firstly compute the latent representations $z_i$ of all samples in the test set.
We use $80\%$ of the samples to train a \textit{domain classifier} using the data-source ID as label. 
We subsequently evaluate the domain-invariance of the representations by calculating the achieved accuracies of these domain classifiers on the remaining $20\%$ of the test set representations.
Note that a lower 4-class domain-classifier accuracy corresponds to a higher domain-invariance of the learned representations, which is favorable for multi-source learning.
We considered a multitude of learning machines as domain classifiers (e.g., support vector machines (SVM), quadratic discriminant analysis). Since all domain classifiers showed similar results we report our results with linear SVMs in Section~\ref{sec:invariance}.

\section{Experimental Results}

\subsection{Emotion Classification}

\autoref{tab:results} summarizes our results on the emotion classification task.
Here \textit{local models} indicate four independent networks processed on each data-source individually (i.e., \textit{DAPE} with $M=1$), which would be a baseline approach if one is not interested in domain-privacy or representation invariance.
To the contrary, the \textit{global model} indicates a baseline where one trains a single, unified model by pooling all data-source inputs into one unified training set, however constraining both the amount of training data and the encoder specifications to be uniform and shared across all data-sources.
DANN~\cite{ganin2016domain} depicts an invariant representation learning approach that was previously considered for EEG decoding in \cite{ozdenizci2020learning,rayatdoost2021subject}, where one can train a model adversarially by censoring the data-source ID relevant information from latent representations to impose invariance, without considering data-source privacy.

While the global model and DANN perform only slightly above chance level (33\%) in emotion classification, we observe that \textit{DAPE} and \textit{aDAPE} clearly outperform the baseline approaches that learns a unified model for emotion classification, while simultaneously ensuring data-source privacy.
Furthermore, we observe that \textit{aDAPE} is also only slightly below the average of the extensive local models (49.09\% vs. 51.91\%), although solving this task across multiple domains poses a more challenging problem than learning locally.

\subsection{Domain-Invariance of the Learned Representations}
\label{sec:invariance}

Domain-invariance of the learned representations estimated through the linear SVM are shown at the bottom row of \autoref{tab:results}.
We observe that \textit{aDAPE} shows the highest domain-invariance in the learned representations, where in all other models the linear SVM was able to deduce the data-source ID from the representations with $>60\%$ accuracy.
Achieved domain-invariance is $\sim14\%$ higher for \textit{aDAPE} than for \textit{DANN}, and $\sim47\%$ higher than for the local models.
Furthermore, it is important to note that the performance of the domain classifier is much higher ($\sim30\%$) for \textit{aDAPE} than for \textit{DAPE}, from an ablation study perspective.
This significant gap results in the use of adaptive batch normalization layers.
Overall emotion classification accuracy of \textit{aDAPE} is higher than of \textit{DAPE}, and achieved domain-invariance of representations is also stronger (while still being slightly above the chance-level), confirming our hypothesis that using domain-invariant representations in multi-source settings helps by making the classification task more robust.

\section{Conclusion}

We present \textit{Domain Aligned Private Encoders} as a multi-source learning framework for domain-invariant representation learning\footnote{\url{https://github.com/philipph77/DAPE-Framework}}, and demonstrate its feasibility on an EEG-based emotion classification task using data from four publicly available datasets.
Different than state-of-the-art adversarial training approaches to learn invariant EEG representations \cite{ozdenizci2020learning,jeon2021mutual,rayatdoost2021subject}, we utilize an MMD based domain alignment loss \cite{jimenez2020custom,chen2021ms} across dataset-specific private feature encoders.
Proposed deep neural signal processing architecture encodes multi-channel EEG time-series signals privately at the source end and only the learned representations are shared to a global classifier.
Our decentralized invariant representation learning approach and use of adaptive batch normalization layers improved performance in our experimental analyses.

Accessible EEG data is generally present in small datasets, which are also distributed across various countries and/or laboratories. 
Due to naturally occurring data privacy concerns, we believe that such architectures as proposed in this work make common use of private data representations more convenient and real-life applicable (e.g., clinical time-series monitoring \cite{ozdenizci2018time}, personalized BCI-based stroke rehabilitation protocols \cite{agarwal2019protecting,mastakouri2017personalized}), hence our work advances the area of cross-domain transfer learning for EEG signals in that sense.

\newpage
\bibliographystyle{IEEE.bst}
\bibliography{mybib}

\end{document}